\documentclass[11pt]{article}

\usepackage[preprint]{acl}
\makeatletter
\renewcommand\outauthor{%
    \begin{tabular}[t]{c}
    \ifacl@anonymize%
        \bfseries Anonymous EMNLP submission
    \else
        \@author%
    \fi
    \end{tabular}}
\makeatother

\usepackage{times}
\usepackage{latexsym}

\usepackage[T1]{fontenc}

\usepackage[utf8]{inputenc}

\usepackage{microtype}
\IfFileExists{inconsolata.sty}{\usepackage{inconsolata}}{}
\usepackage{booktabs}
\usepackage{array}
\usepackage{multirow}
\usepackage{colortbl}
\usepackage{xcolor}
\usepackage{subcaption}
\usepackage{placeins}
\usepackage{float}
\usepackage{amsmath}
\usepackage{amssymb}
\usepackage{algorithm}
\usepackage{algpseudocode}
\usepackage{graphicx}
\graphicspath{{images/}}
\usepackage{hyperref}
\usepackage{url}
\usepackage[breakable]{tcolorbox}

\floatstyle{ruled}
\restylefloat{algorithm}

\newcommand{\nickname}{Infini Memory}
\newcommand{\jsonquote}{\char34}

\definecolor{promptblue}{RGB}{248,250,252}
\definecolor{promptborder}{RGB}{96,165,250}
\definecolor{prompttitle}{RGB}{37,99,235}

\newtcolorbox{promptbox}[1][]{
  colback=promptblue,
  colframe=promptborder,
  colbacktitle=prompttitle,
  coltitle=white,
  fonttitle=\bfseries\small,
  title=#1,
  boxrule=0.55pt,
  titlerule=0pt,
  arc=2pt,
  left=7pt,
  right=7pt,
  top=5pt,
  bottom=5pt,
  toptitle=3pt,
  bottomtitle=3pt,
  before skip=0.75\baselineskip,
  after skip=0.75\baselineskip,
  breakable,
}

\newtcolorbox{promptboxwide}[1][]{
  colback=promptblue,
  colframe=promptborder,
  colbacktitle=prompttitle,
  coltitle=white,
  fonttitle=\bfseries\small,
  title=#1,
  boxrule=0.55pt,
  titlerule=0pt,
  arc=2pt,
  left=8pt,
  right=8pt,
  top=5pt,
  bottom=5pt,
  toptitle=3pt,
  bottomtitle=3pt,
  before skip=0.8\baselineskip,
  after skip=0.8\baselineskip,
  breakable,
}

\newenvironment{widepromptbox}[1][]{
  \begin{promptboxwide}[#1]
  \ttfamily\footnotesize
  \setlength{\parindent}{0pt}
  \setlength{\parskip}{1pt}
  \linespread{0.96}\selectfont
}{
  \end{promptboxwide}
}

\hypersetup{hidelinks}
\title{\nickname{}: Maintainable Topic Documents for Long-Term LLM Agent Memory}

\author{
  Suozhao Ji$^{1}$ \quad
  Baodong Wu$^{1,\ast}$ \quad
  Zehao Wang$^{2}$ \quad
  Lei Xia$^{1}$ \quad
  Qingping Li$^{1}$ \\
  Ruisong Wang$^{1}$ \quad
  Wenbo Ding$^{2}$ \quad
  Zhenhua Zhu$^{2}$ \quad
  Boxun Li$^{1}$ \quad
  Guohao Dai$^{3,1}$ \quad
  Yu Wang$^{2,\ast}$ \\[4pt]
  $^{1}$Infinigence AI \quad
  $^{2}$Tsinghua University \quad
  $^{3}$Shanghai Jiaotong University \\[2pt]
  \texttt{\{jisuozhao, wubaodong, xialei, liqingping, wangruisong, liboxun\}@infini-ai.com} \\
  \texttt{\{wangzeha24@mails, ding.wenbo@sz, zhuzhenhua@mail, yu-wang@mail\}.tsinghua.edu.cn} \\
  \texttt{daiguohao@sjtu.edu.cn} \\[2pt]
  $^{\ast}$Corresponding authors \\[2pt]
  \textbf{Code:} \url{https://github.com/infinigence/Infini-Memory}
}
\date{}

\begin{document}
\maketitle

\begin{abstract}
Long-term LLM agents need persistent memory that can track changing facts and provide relevant evidence across sessions. Existing memory systems often store observations as isolated records, summaries, or indexed fragments, which makes evidence aggregation, fact revision, and memory maintenance difficult. We propose \textbf{\nickname{}}, a maintainable text-based persistent memory architecture that treats agent memory as topic-structured documents. Each topic document serves as a semantic unit for collecting related evidence, preserving metadata, and revising facts over time. New observations are first staged in a buffer and periodically consolidated into coherent textual contexts. At inference time, an agentic retrieval procedure lets the LLM read memory through iterative tool calls rather than a single retrieval step. On MemoryAgentBench, \nickname{} achieves 64.7\% overall score. Ablations show that topic-structured maintenance and iterative evidence inspection improve complementary aspects of long-term memory use.
\end{abstract}

\section{Introduction}
LLM agents increasingly operate over long horizon across many sessions, but an LLM's context window only bounds how much input the model can attend to in a single forward pass. Increasing its length exposes more history to the model, but does not by itself specify which information should be retained, how stale information should be revised, or how related evidence should be organized for future use~\cite{packer2023memgpt, sumers2023coala}.

To address this gap, recent agent systems introduce memory mechanisms that retain information beyond the immediate prompt, including prior interactions, tool observations, task facts, and user preferences~\cite{yao2023react, park2023generativeagents, shinn2023reflexion}. Memory has been studied as a component of language-agent architectures and long-term conversational agents~\cite{zhong2024memorybank, chhikara2025mem0, xu2025amem, sumers2023coala}, with a parallel line of work introducing benchmarks for multi-turn memory capabilities~\cite{maharana2024locomo, wu2025longmemeval, hu2025memoryagentbench}. In this paper, we use \textit{persistent memory} to refer to an external, editable memory state that survives across interaction sessions and can be written, updated, and queried by an agent during inference.



\begin{figure}[t]
\centering
\includegraphics[width=\columnwidth]{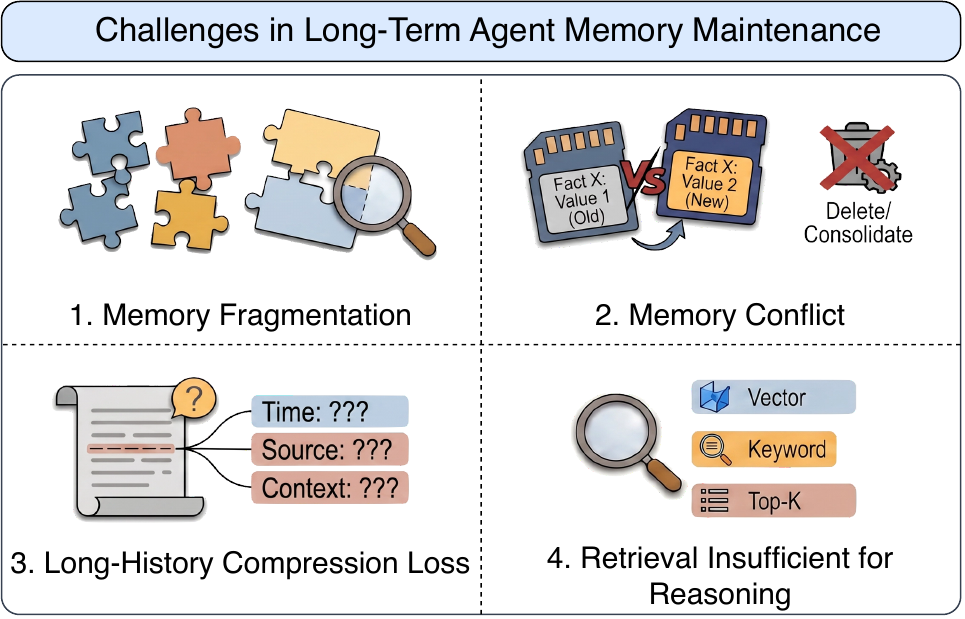}
\caption{Four recurring challenges in long-term agent memory maintenance: (1)~related evidence is scattered across isolated records; (2)~old and new versions of the same fact coexist without reconciliation; (3)~summarization loses temporal, source, and contextual cues; (4)~single-shot retrieval returns fragments insufficient for multi-hop reasoning.}\label{fig:challenges}
\end{figure}

These memory systems typically maintain information outside the model and retrieve relevant content back into the prompt at inference time. Existing designs include text memories with summary-based dialogue memories~\cite{tan2025prospect}, vector retrieval~\cite{park2023generativeagents, chhikara2025mem0}, hierarchical memory managers~\cite{packer2023memgpt, kang2025memory, fang2026lightmem}, and knowledge-graph memory layers~\cite{rasmussen2025zep, gutierrez2025hipporag2, shu2026remem}.
These systems demonstrate the usefulness of external memory, but they also exhibit four recurring failure modes (Figure~\ref{fig:challenges}). When memory is represented mainly as independent retrievable items, evidence about the same user, task, or event can be distributed across many small records (\textit{memory fragmentation}). When newer observations contradict older ones, append-style storage can leave both versions available for retrieval (\textit{memory conflict}). When long histories are compressed into summaries, temporal order and source cues may be weakened (\textit{compression loss}). Standard retrieval methods based on vector similarity, keyword matching, or fixed top-$k$ procedures may return isolated fragments rather than enough evidence for temporally grounded reasoning (\textit{insufficient retrieval}).
We use \textit{fragments} to refer to such retrieved pieces: evidence that may be relevant to a query but lacks the local context needed to resolve it.

These observations motivate treating persistent memory as a lifecycle maintenance problem. A memory module for long-term agents should support three coupled operations. First, it should \textit{write} durable information from interactions into the memory state. Second, it should \textit{maintain} that state by grouping related evidence so that facts about the same user, task, or event are no longer scattered, reconciling new observations against contradicted older ones, and preserving temporal and source cues even as content is condensed. Third, it should \textit{read} memory by retrieving evidence with sufficient local context rather than isolated fragments.

We propose \nickname{}, a maintainable persistent memory architecture that represents external memory as topic-structured documents. Each topic document groups related evidence under a shared subject and carries entry-level metadata that retains temporal and source cues as content is rewritten. The system separates frequent writes from less frequent structural consolidation: new memory candidates are first appended to a buffer document, then periodically rewritten, split, updated, and merged into topic documents. At query time, the agent retrieves from this document library and expands local context around matching evidence. Retrieval over this library can use lexical indexing over plain-text documents rather than a vector or graph database backend.

The main contributions of this work are as follows:
\begin{itemize}
    \item We introduce a document-based persistent memory architecture for long-term LLM agents. It organizes memory as topic documents and maintains them through buffered writing and periodic consolidation, avoiding a mandatory dependency on vector or graph databases.

    \item We present an agentic retrieval strategy in which the LLM controls a multi-step search process over structured memory tools. The strategy supports iterative evidence inspection, local context expansion, and answer-oriented evidence assembly.

    \item We evaluate \nickname{} on MemoryAgentBench, where its agentic retrieval variant achieves 64.7\% overall, and use controlled variants to analyze maintenance and retrieval choices.
\end{itemize}

\section{Related Work}

\subsection{Persistent Memory Representation and Maintenance}

Recent work on LLM agents has increasingly treated memory as an external state that must be written, updated, and retrieved across interactions. Early persistent-memory systems mainly extend the effective context available to an LLM\@. MemGPT introduces virtual context management, in which an agent moves information between limited in-context memory and external storage through explicit control operations~\cite{packer2023memgpt}. MemoryBank stores long-term user memories and updates them over time with mechanisms inspired by the Ebbinghaus forgetting curve~\cite{zhong2024memorybank}. Mem0 further develops this line by dynamically extracting, consolidating, and retrieving salient information from conversations, with a graph-based variant for relational structure~\cite{chhikara2025mem0}. These systems show that persistent memory is useful for long-term interaction, but they often store memory as compact entries, summaries, or indexed fragments, which can make later revision and evidence reconstruction difficult when information is distributed across many interactions.

A second line of work focuses on structured memory organization. Graph-based and associative approaches, such as HippoRAG-v2, use non-parametric memory structures to support factual, associative, and sense-making retrieval~\cite{gutierrez2025hipporag2}. REMem represents episodic memory with a hybrid graph over time-aware gists and facts, targeting recollection and reasoning over event histories~\cite{shu2026remem}. A-MEM proposes an agentic memory system inspired by the Zettelkasten method, where each memory is stored as an atomic note with structured attributes and dynamically generated links to related memories~\cite{xu2025amem}. These methods improve memory organization beyond flat retrieval, but they often rely on atomic notes, embeddings, or graph structures as the main substrate.


LightMem is also relevant because it separates memory processing into stages: sensory filtering, topic-aware short-term consolidation, and offline long-term update~\cite{fang2026lightmem}.\ \nickname{}\ shares the motivation of reducing online maintenance overhead, but differs in emphasizing plain-text topic documents and explicit consolidation operations. This makes its design more aligned with systems where interpretability, editable state, and infrastructure simplicity are important.

\subsection{Retrieval and Evaluation for Long-Term Memory Agents}

Retrieval is a key difficulty for long-term memory because relevant evidence may be scattered across many interactions. Standard retrieval pipelines usually rely on vector similarity, keyword matching, or a fixed top-$k$ procedure. These methods are efficient, but they may return isolated fragments rather than enough evidence for temporally grounded reasoning or contradiction resolution. Recent systems therefore move toward more active retrieval procedures. REMem, for example, uses an agentic retriever with curated tools to iteratively retrieve and reason over episodic memory graphs~\cite{shu2026remem}. A-MEM introduces agency mainly in memory construction and organization, dynamically creating notes, attributes, and links when new memories arrive~\cite{xu2025amem}.\ \nickname{}\ extends this direction to the read path over structured text memory: the LLM can iteratively choose memory tools, inspect intermediate results, expand local context, and assemble evidence before answering.

Benchmarks for long-term memory have also shifted from static long-context understanding toward interactive memory evaluation. LoCoMo evaluates very long-term conversational memory over multi-session dialogues with question answering, event summarization, and multimodal dialogue generation tasks~\cite{maharana2024locomo}. LongMemEval focuses on long-term interactive memory for chat assistants and evaluates information extraction, multi-session reasoning, temporal reasoning, knowledge updates, and abstention~\cite{wu2025longmemeval}. These benchmarks are useful for testing long-context recall and temporally grounded dialogue understanding, but they do not fully isolate the operational abilities required by memory agents that incrementally store, revise, and retrieve information.

MemoryAgentBench is more directly aligned with the goals of this work. It evaluates memory agents through incremental multi-turn interactions and identifies four core competencies: accurate retrieval, test-time learning, long-range understanding, and selective forgetting~\cite{hu2025memoryagentbench}. These competencies match the main design questions addressed by \nickname{}: whether the system can retrieve relevant evidence, acquire new information during deployment, integrate long-range context, and revise outdated memory. We therefore use MemoryAgentBench as the main evaluation setting, while interpreting results as benchmark-level evidence rather than as a complete characterization of all long-term memory use cases.

\section{Memory Design}

\subsection{\nickname{} Design Overview}

\nickname{} represents persistent memory as a library of \emph{topic documents}, where a topic denotes a maintenance scope that groups entries handled together by later operations (routing, splitting, merging). This scope is defined by how the memory will be used in future interactions. For example, entries about a stable user preference or an ongoing project may form a topic because they provide context for the same class of future questions and updates. This design avoids two less desirable extremes. (1) If memory is stored as isolated records, related evidence may be separated and later retrieval may return fragments without enough context; (2) if all memory is stored in a single chronological log, later operations may need to scan or rewrite unrelated history. We discuss the alternatives we rejected (pure vector store, pure knowledge graph, no buffer) in Appendix~\ref{appendix:rationale}.
Topic documents provide a bounded unit that can preserve related entries and their metadata under local headings, while keeping each document focused enough for rewriting, splitting, and merging.

Based on topic documents, \nickname{} organizes its memory pipeline around representation, writing, consolidation, and retrieval. Each topic document stores a summary, a hierarchical body, and entry-level metadata to organize related evidence within a bounded document scope (Section~\ref{sec:memory-representation}). New memories first enter a short-term buffer named \texttt{CURRENT}, so frequent writes do not repeatedly rewrite the topic library; consolidation is triggered after enough related evidence accumulates (Section~\ref{sec:memory-writing}). At inference time, the LLM iteratively searches, inspects, and expands context from the maintained topic library to recover evidence beyond a single retrieval result (Section~\ref{sec:memory-retrieval}). The structured text backend keeps the default system self-contained and leaves room for optional retrieval or maintenance extensions (Section~\ref{sec:deployment-extensibility}).

\begin{figure}[t]
\centering
\includegraphics[width=\columnwidth]{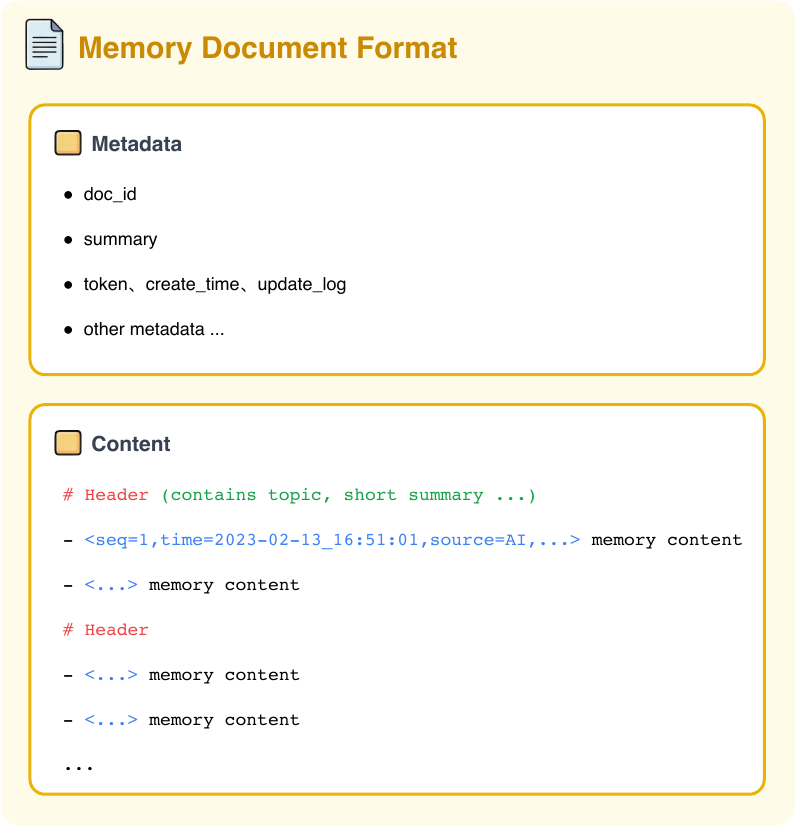}
\caption{Topic document format used by \nickname{}.}\label{fig:doc-format}
\end{figure}

\subsection{Topic Document Representation}\label{sec:memory-representation}

\nickname{} stores persistent memory as topic documents (Figure~\ref{fig:doc-format}), where each document groups related facts, preferences, and event cues under a shared topic. A document contains a metadata header, \texttt{\{id, summary, token\_count, created\_time, update\_log, aux\}}, and a hierarchical body. The body uses topic and subtopic headings to organize unordered-list memory entries, each prefixed with a parsable signature \texttt{<seq=\ldots, time=\ldots, source=\ldots>}. This representation preserves local context while keeping temporal order, provenance, and later revision operations explicit.

\paragraph{Entry-level metadata.} Each memory entry carries at least a sequence number \texttt{seq}, which increases monotonically with each write call. When temporal information is available from the interaction content, a \texttt{time} field is recorded. When information is extracted from the model response, a \texttt{source=AI} tag can be attached. The metadata signature can be extended with domain-specific fields such as entity type, namespace, machine identifier, IP address, or sensitivity level.

This storage format has three practical benefits. First, the document summary and body can be refreshed together when a topic document is maintained, keeping retrieval metadata aligned with the underlying evidence. Second, temporal and source cues move with each entry when content is modified, e.g., rewritten, split, or merged. Third, each memory entry carries metadata that makes later revision explicit. When new evidence updates earlier content, \texttt{seq} and \texttt{time} provide ordering cues for superseding outdated entries or applying explicit deletion rules.

\subsection{Buffered Writing and Consolidation}\label{sec:memory-writing}

\begin{figure}[t]
\centering
\includegraphics[width=\columnwidth]{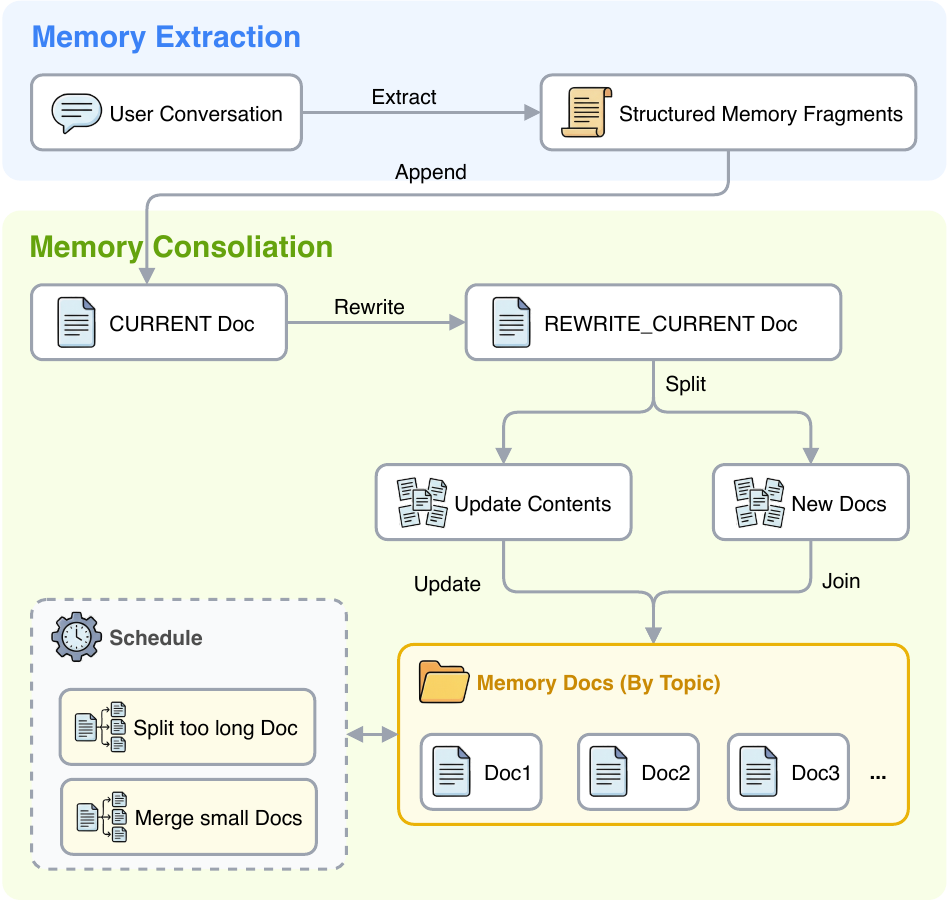}
\caption{Memory writing and consolidation pipeline.}\label{fig:write-pipeline}
\end{figure}

The writing and consolidation pipeline (Figure~\ref{fig:write-pipeline}; full procedure in Algorithm~\ref{alg:write-consolidate}) is designed around the mismatch between the frequency of memory extraction and the scope of memory maintenance. Memory extraction may happen after every interaction, but consolidation should not. Updating the topic library after each extracted entry would require repeated topic routing, contradiction checking, and document rewriting. Such eager maintenance may also introduce unstable edits before enough related evidence is available. \nickname{} therefore introduces \texttt{CURRENT} as a buffer for recent entries.

The \texttt{CURRENT} buffer collects extracted memories in append form. Appending to this buffer does not require scanning or rewriting existing topic documents. More importantly, the buffer preserves the short-range coherence of recent interactions. Several adjacent turns often describe the same task, correct the same fact, or refine the same preference. Keeping them together before consolidation allows the system to resolve local redundancy and contradictions before they enter the topic library.

The buffer is flushed when it reaches a token threshold or remains active for a predefined time window:

\begin{equation}
\mathrm{flush}(C) =
\bigl(|C| \geq \tau_{\mathrm{tok}}\bigr)
\lor
\bigl(\Delta t(C) \geq \tau_{\mathrm{time}}\bigr).
\label{eq:flush-trigger}
\end{equation}
where \(C\) denotes the current buffer, \(\tau_{\mathrm{tok}}\) is a token threshold, and \(\tau_{\mathrm{time}}\) is a time threshold.

When the buffer is flushed, the system rewrites \texttt{CURRENT} into \texttt{REWRITE\_CURRENT}. This intermediate draft is not a persistent memory store. It is a normalized view of the recent buffer, created to make library update easier. The rewrite step groups locally related entries, removes redundant statements, preserves useful metadata, and marks possible updates to earlier facts. For example, several adjacent entries may be merged into one statement with a sequence range, while a correction may be marked as superseding an earlier entry. The full prompt invariants enforced at this stage are listed in Appendix~\ref{appendix:prompts}.

The normalized draft is then routed into the topic library. For each block in \texttt{REWRITE\_CURRENT}, the planner decides whether it should update an existing topic document or create a new one. If the block extends an existing topic, it is inserted into the relevant document region. If it changes an earlier fact, the planner records the update relation and rewrites the affected local context. If it does not fit any existing maintenance scope, a new topic document is created. This step combines topic assignment and revision, because the correct target document is the one in which the block can be maintained together with related evidence. This procedure is given in Algorithm~\ref{alg:safe-plan}.

After the update is applied, \texttt{CURRENT} is cleared for the next writing interval. Recent buffer content remains available to the retrieval module before consolidation, so newly written information can still be used in answers. This avoids a gap between extraction and retrievability.

The topic library is periodically updated through split and merge operations. Overgrown documents are split to reduce overly broad local context, while fragmented documents are merged when they describe the same maintenance scope. After each structural update, summaries and metadata are refreshed to support future routing and retrieval.

\subsection{Agentic Retrieval over Topic Documents}\label{sec:memory-retrieval}

\begin{figure}[t]
\centering
\includegraphics[width=\columnwidth]{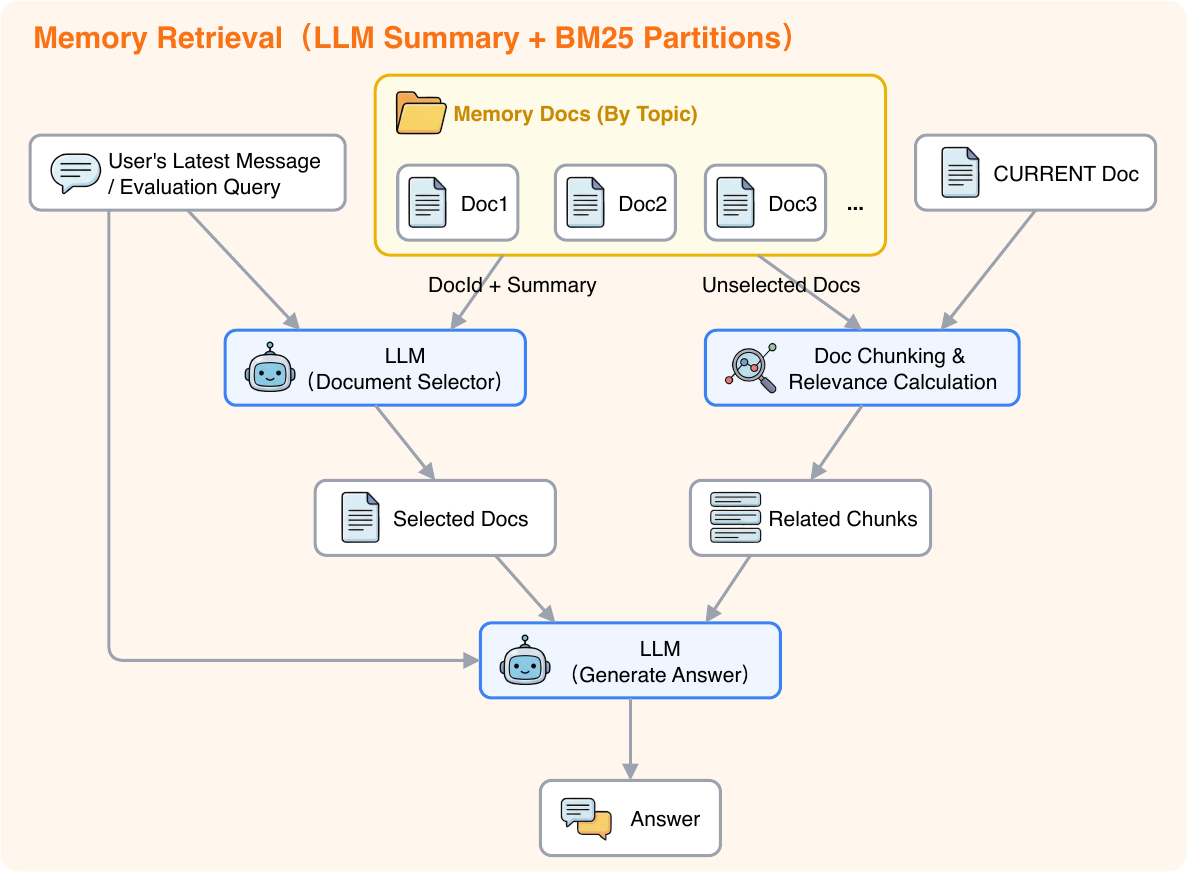}
\caption{Hybrid retrieval variant (LLM Summary + BM25 Partitions). The LLM selects candidate documents by summary, and BM25 supplements with lexically matched partitions.}\label{fig:retrieval}
\end{figure}

\begin{figure}[t]
\centering
\includegraphics[width=\columnwidth]{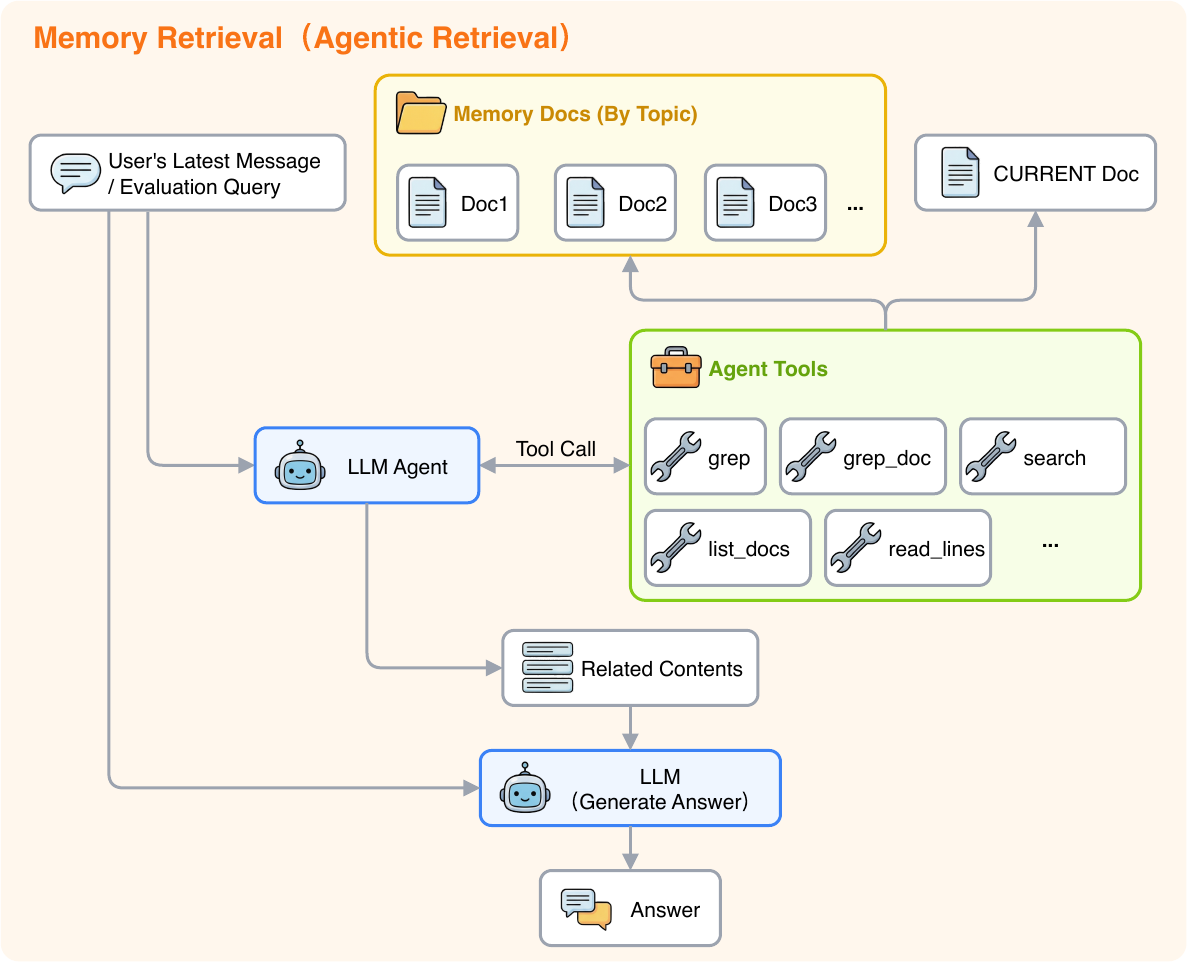}
\caption{Agentic retrieval variant. The LLM agent iteratively calls memory tools (\texttt{grep}, \texttt{grep\_doc}, \texttt{search}, \texttt{list\_docs}, \texttt{read\_lines}) to search, verify, and expand evidence across topic documents and the \texttt{CURRENT} buffer before generating the final answer.}\label{fig:retrieval-agentic}
\end{figure}

The retrieval module is responsible for turning the topic library into evidence for answer generation. \nickname{} supports two retrieval variants: a hybrid reader (\nickname{}-H) that combines LLM-based summary selection with BM25 partition retrieval (Figure~\ref{fig:retrieval}), and an agentic reader (\nickname{}-A)  in which the LLM controls a multi-step search process over memory tools (Figure~\ref{fig:retrieval-agentic}). In the agentic variant, the model selects which tool to call, inspects intermediate results, expands local context when needed, and decides when the collected evidence is sufficient. It is a tool-guided retrieval workflow built on top of topic documents.

Long-term memory questions often require more than isolated snippets. They may depend on related events, updated facts, or surrounding context. A single-shot retrieval can miss these connections. With topic documents, a matched entry can be expanded into its local block, where temporal and source metadata help the system identify the relevant evidence.

At the start of retrieval, the system exposes a document catalog and a set of memory tools. The catalog contains document identifiers and summaries. For small libraries, the catalog can be provided directly. For larger libraries, the agent can inspect the catalog through paging or search. The default tools include global lexical search, document-local pattern search, catalog inspection, and line-range reading. These tools correspond to different retrieval behaviors: broad search finds candidate regions, local search verifies precise matches, and line-range reading recovers the context around evidence.

During retrieval, the agent alternates between tool calls and evidence inspection. Early steps usually identify candidate documents or headings. Later steps read local regions and check whether the evidence supports the query. The loop stops when the agent returns a stop decision, reaches a maximum number of iterations, reaches an evidence budget, or fails to obtain new useful evidence. These limits are important because agentic retrieval increases test-time computation compared with single-shot retrieval. The full retrieval loop, including the BM25 fallback path, is given in Algorithm~\ref{alg:retrieval}; the prompt that drives the per-step search behavior appears in Appendix~\ref{appendix:prompts}.

The final evidence set may contain document-level selections, heading-level blocks, or expanded line ranges. Snippet-level results are expanded to the nearest coherent heading block when possible. The final context also includes recent entries from \texttt{CURRENT}, so unconsolidated memories remain accessible. If the agent returns no evidence or too little evidence, the system runs a conservative lexical fallback over the topic library. This fallback is used as a recall guard and does not replace the agentic retrieval policy.

\subsection{Deployment and Extensibility}\label{sec:deployment-extensibility}
\nickname{} uses structured text as the default memory carrier. This choice keeps the default backend simple because the system can operate with ordinary document storage, lexical indexing, and deterministic file inspection tools. It does not require configuring a vector database, graph database, or external memory service before the system can run.

This design should be interpreted as backend-light rather than computation-free. Agentic retrieval may use more tool calls or LLM tokens than a single retrieval step. Periodic consolidation also introduces maintenance cost. The intended benefit is that the default memory state remains readable, editable, and portable, while more specialized retrieval backends can be added when needed.

The same abstraction can support domain-specific extensions. The metadata area of each memory entry can store namespaces, entity schemas, access rules, or retention policies. Additional tools can expose vector search, graph traversal, database lookup, or permission checks. These extensions can be integrated as retrieval tools or maintenance rules while preserving topic documents as the shared memory state.
\newsavebox{\mainresultsbox}
\begin{table*}[t]
\centering
\footnotesize
\setlength{\tabcolsep}{1.0pt}
\renewcommand{\arraystretch}{1.08}
\begin{lrbox}{\mainresultsbox}
\begin{tabular}{@{}l*{15}{c}@{}}
\toprule
\multicolumn{1}{c}{Method} & \multicolumn{5}{c}{AR} & \multicolumn{3}{c}{TTL} & \multicolumn{3}{c}{LRU} & \multicolumn{3}{c}{SF} & Overall Score \\
\cmidrule(lr){2-6} \cmidrule(lr){7-9} \cmidrule(lr){10-12} \cmidrule(lr){13-15}
 & SH-QA & MH-QA & LME & Event & Avg. & MCC & Rec. & Avg. & Summ. & DetQA & Avg. & FC-SH & FC-MH & Avg. & \\
\midrule
\multicolumn{16}{@{}l}{\textit{Baselines}} \\
RAPTOR & 32.0 & 39.0 & 38.2 & 48.2 & 39.4 & 56.8 & 14.0 & 35.4 & 25.3 & 45.6 & 35.5 & 16.0 & 2.0 & 9.0 & 29.8 \\
MemoRAG & 35.0 & 36.0 & 25.0 & 54.4 & 37.6 & 75.0 & 15.7 & 45.4 & 15.3 & 53.5 & 34.4 & 25.0 & 8.0 & 16.5 & 33.5 \\
HippoRAG-v2 & 78.0 & 70.0 & 54.7 & 72.2 & 68.7 & 65.2 & 12.3 & 38.8 & 32.1 & 54.3 & 43.2 & 58.0 & 5.0 & 31.5 & 45.5 \\
Mem0 & 28.0 & 36.0 & 42.0 & 35.5 & 35.4 & 35.2 & 11.2 & 23.2 & 12.4 & 42.3 & 27.4 & 25.0 & 3.0 & 14.0 & 25.0 \\
MemGPT & 49.0 & 30.0 & 42.7 & 45.6 & 41.8 & 65.6 & 14.3 & 40.0 & 10.9 & 40.2 & 25.6 & 32.0 & 3.0 & 17.5 & 31.2 \\
LightMem & 57.0 & 34.0 & 69.2 & 39.2 & 49.9 & 58.0 & 12.7 & 35.4 & 27.5 & 41.0 & 34.2 & 29.0 & 6.0 & 17.5 & 34.2 \\
REMem & 62.0 & 42.0 & 62.3 & 41.2 & 51.9 & 80.0 & 13.2 & 46.6 & 23.4 & 47.6 & 35.5 & 32.0 & 9.0 & 20.5 & 38.6 \\
\midrule
\multicolumn{16}{@{}l}{\textit{\nickname{} variants}} \\
\nickname{}-H & 77.0 & 76.0 & 76.0 & 83.0 & 78.0 & 81.0 & 15.8 & 48.4 & 57.2 & \textbf{78.4} & 67.8 & 80.0 & 22.0 & 51.0 & 61.3 \\
\textbf{\nickname{}-A} & \textbf{83.0} & \textbf{79.0} & \textbf{79.3} & \textbf{83.6} & \textbf{81.2} & \textbf{84.0} & \textbf{18.0} & \textbf{51.0} & \textbf{59.9} & 77.2 & \textbf{68.6} & \textbf{81.0} & \textbf{35.0} & \textbf{58.0} & \textbf{64.7} \\
\bottomrule
\end{tabular}
\end{lrbox}
\usebox{\mainresultsbox}\par
\vspace{0.25em}
\begin{minipage}{\wd\mainresultsbox}
\footnotesize
\raggedright
\vspace{0.25em}
\textit{Note.} AR = Accurate Retrieval; TTL = Test-Time Learning; LRU = Long-Range Understanding; SF = Selective Forgetting. Sub-datasets (scores are accuracy unless otherwise noted): SH-QA / MH-QA = single-/multi-hop document QA; LME = LongMemEval (S$^\star$), a reconstructed multi-session dialogue variant; Event = EventQA, temporal-event reasoning over long narratives; MCC = multi-class classification; Rec. = movie recommendation (Recall@5); Summ. = novel summarization (Fluency $\times$ F1); DetQA = detective reasoning QA; FC-SH / FC-MH = FactConsolidation single-/multi-hop selective forgetting. Bold marks the best value in each column.
\end{minipage}
\caption{Full MemoryAgentBench results. All methods use \texttt{gpt-5-mini} as the backbone model with an input chunk size of 4096 tokens.}\label{tab:main-results}
\end{table*}

\section{Experiments}
\subsection{Experimental Setup}\label{sec:setup}

\paragraph{Benchmark and Models.}
We evaluate our method on MemoryAgentBench~\cite{hu2025memoryagentbench}, a benchmark for external memory mechanisms across four capabilities: Accurate Retrieval (AR) for factual recall from long histories, Test-Time Learning (TTL) for in-context rule acquisition, Long-Range Understanding (LRU) for extended narrative comprehension, and Selective Forgetting (SF) for updating outdated information.

We use \texttt{gpt-5-mini} as the base model. System outputs are evaluated with an LLM-as-Judge protocol using \texttt{gpt-5}, where the judge assigns a binary correctness judgment based on the question, reference answer, and model output.

\paragraph{Baselines.}
We compare \nickname{} against seven memory baselines. We use the official integrations supplied with MemoryAgentBench for RAPTOR~\cite{sarthi2024raptor}, MemoRAG~\cite{qian2025memorag}, HippoRAG-v2~\cite{gutierrez2025hipporag2}, Mem0~\cite{chhikara2025mem0}, MemGPT~\cite{packer2023memgpt}, LightMem~\cite{fang2026lightmem}, and REMem~\cite{shu2026remem}. All baselines follow the same 4096-token chunking strategy as the benchmark's standard configuration. Each baseline retains its own retrieval and indexing logic under its official integration.

\paragraph{Memory and retrieval configuration.}
For memory maintenance, \nickname{} triggers a \texttt{CURRENT} buffer rewrite upon reaching either 5000 tokens or a time threshold; it splits topic documents exceeding 5000 tokens and merges documents under 1000 tokens based on summary similarity.
During retrieval, the LLM receives the query alongside an \texttt{id + summary} catalog. The agent can invoke memory tools for up to seven rounds—covering corpus search, regex matching, catalog browsing, and line reading. If agentic retrieval yields insufficient evidence, a BM25-based retriever supplements non-duplicate candidates.

\subsection{Main Benchmark Results}
As shown in Table~\ref{tab:main-results}, \textbf{\nickname{}-A} achieves the highest overall score of 64.7\%, improving over the strongest baseline by 19.2\% points and leading on all four MemoryAgentBench capabilities, with gains of +12.5\% on AR, +4.4\% on TTL, +25.4\% on LRU, and +26.5\% on SF. Compared with the hybrid variant \nickname{}-H (61.3\%), agentic retrieval adds 3.4\% points on average; the largest gain falls on Selective Forgetting (+7.0\%), where explicit temporal and source cues help the reader track revised facts.

The pattern within Long-Range Understanding is more nuanced: the agentic reader leads on summary-oriented questions while the hybrid reader leads on detailed QA, reflecting a trade-off between targeted inspection and broader partition-level coverage.

\paragraph{Discussion: multi-hop selective forgetting (FC-MH).}
The 81.0\%/35.0\% gap between FC-SH and FC-MH reveals a structural limit of write-time consolidation. Single-hop forgetting is settled at write time: the rewrite stage applies recency overrides within a topic document, so the latest version supersedes earlier ones before retrieval. Multi-hop forgetting cannot be settled this way, because a reasoning chain often spans several topic documents updated at different times. No single rewrite pass enforces cross-document consistency, and missing any hop causes cascade failure. This difficulty is intrinsic to the task: the MemoryAgentBench authors report that even o4-mini drops from 80.0\% to 14.0\% on FC-MH as context grows from 6K to 32K tokens~\cite{hu2025memoryagentbench}.

\subsection{Ablation Experiments}\label{sec:ablations}

\begin{table*}[t]
\centering
\small
\begin{minipage}[t]{0.58\textwidth}
\centering
\setlength{\tabcolsep}{5pt}
\renewcommand{\arraystretch}{1.05}
\begin{tabular}{@{}l c l c r@{\,}l@{}}
\toprule
Variant & Maint. & Retrieval & Acc. & \multicolumn{2}{c}{$\Delta$} \\
\midrule
\nickname{}-A (full)   & \checkmark{} & Agentic       & \textbf{79.3} & \multicolumn{2}{c}{---} \\
\midrule
\multicolumn{6}{@{}l}{\textit{Retrieval ablation (maintenance fixed)}} \\
\nickname{}-H (hybrid) & \checkmark{} & Summary+BM25 & 76.0 & $-$&3.3 \\
Summary-only           & \checkmark{} & Summary       & 41.7 & $-$&37.6 \\
\midrule
\multicolumn{6}{@{}l}{\textit{Maintenance ablation (retrieval fixed at hybrid)}} \\
w/o Split \& Merge     &              & Summary+BM25 & 69.3 & $-$&10.0 \\
\bottomrule
\end{tabular}
\captionof{table}{Component ablation on LongMemEval (S$^\star$). \textit{Maint.}: plan-driven split/update and small-document merging. $\Delta$ is the absolute accuracy drop relative to \nickname{}-A.}\label{tab:ablation}
\end{minipage}\hfill
\begin{minipage}[t]{0.38\textwidth}
\centering
\setlength{\tabcolsep}{6pt}
\renewcommand{\arraystretch}{1.05}
\begin{tabular}{@{}r r c r@{\,}l@{}}
\toprule
Threshold & \#Docs & Acc. & \multicolumn{2}{c}{$\Delta$} \\
\midrule
$\geq$1000 & 762 & 74.0 & $-$&2.0 \\
$\geq$3000 & 319 & 74.3 & $-$&1.7 \\
\rowcolor{gray!10}
$\geq$5000$^\dagger$ & 322 & \textbf{76.0} & \multicolumn{2}{c}{---} \\
$\geq$7000 & 280 & 70.7 & $-$&5.3 \\
$\geq$9000 & 255 & 69.7 & $-$&6.3 \\
\bottomrule
\end{tabular}
\captionof{table}{Split-threshold sensitivity (in tokens) on LongMemEval (S$^\star$) with the read path fixed at \nickname{}-H. $^\dagger$ denotes our default; $\Delta$ is the accuracy gap to the default.}\label{tab:split-threshold}
\end{minipage}
\end{table*}

\subsubsection{Structural Maintenance Ablation}
We hold the read path fixed at the hybrid reader (\nickname{}-H, Section~\ref{sec:memory-retrieval}) and disable structural maintenance: no plan-driven split/update and no small-document merging, while retaining append-only writes and the \texttt{CURRENT} rewrite. Accuracy on LongMemEval (S$^\star$) falls from 76.0\% to 69.3\%, a drop of 6.7 points (Table~\ref{tab:ablation}).

\begin{figure}[t]
\centering
\includegraphics[width=\columnwidth]{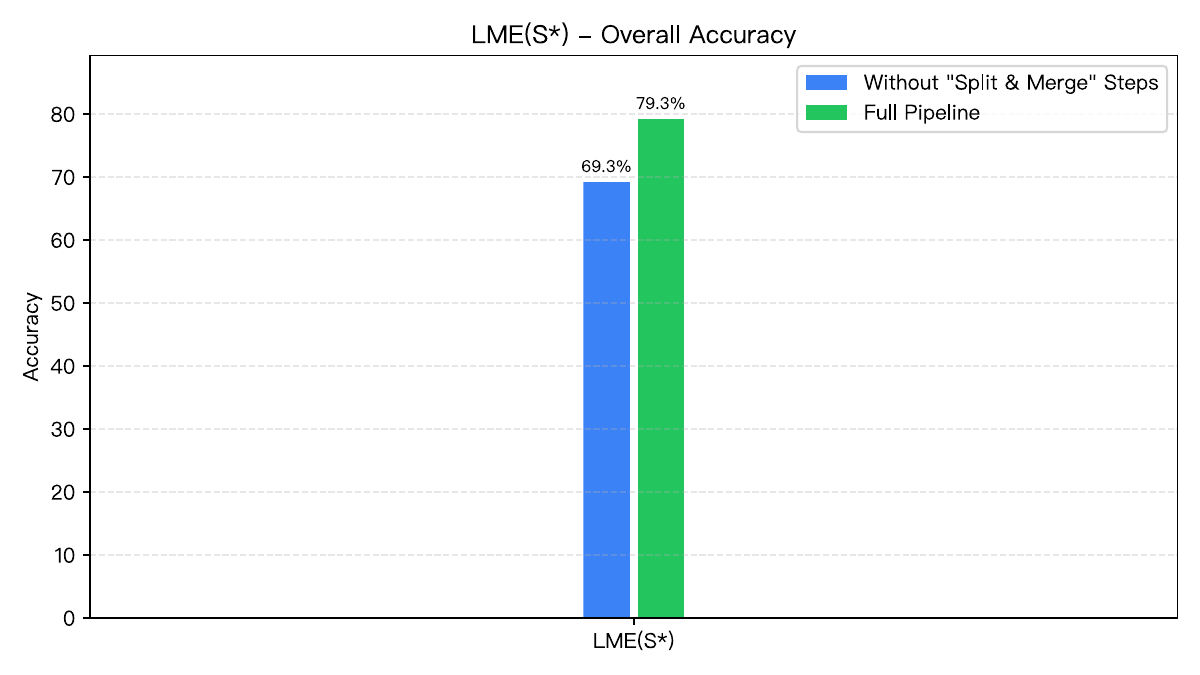}
\vspace{0.4em}
\includegraphics[width=\columnwidth]{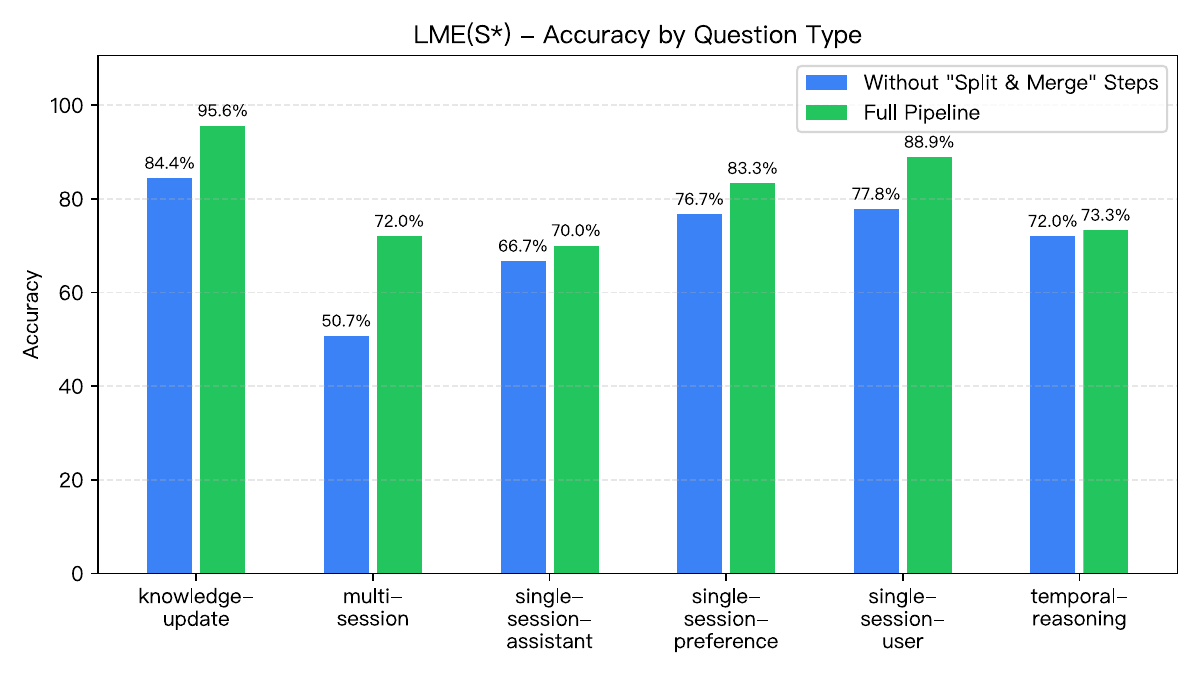}
\caption{Ablation results after removing structural maintenance. Top: overall accuracy on LongMemEval (S$^\star$). Bottom: accuracy by question type.}\label{fig:ablation-structure}
\end{figure}

Figure~\ref{fig:ablation-structure} breaks this gap down by question type. The shortfall concentrates on knowledge-update and multi-session questions, both of which depend on evidence drawn from distant turns and reconciled at read time. Without split and merge, related facts remain in whichever document they were first appended to, and superseded entries continue to coexist with their replacements, so the reader can no longer assemble a coherent and up-to-date answer across the relevant turns. Question types that resolve within a single session degrade much less, since the \texttt{CURRENT} rewrite alone already removes local duplicates and contradictions when the supporting evidence is nearby.

\subsubsection{Retrieval Strategy Ablation}
\begin{figure}[t]
\centering
\includegraphics[width=\columnwidth]{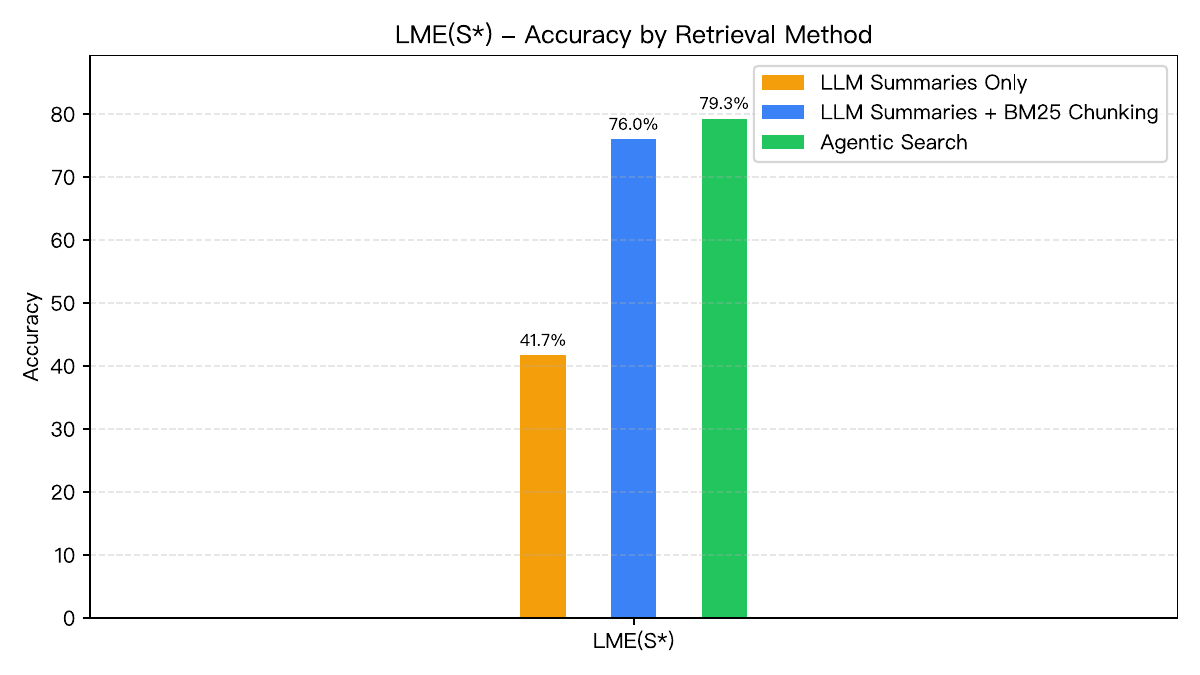}
\caption{Retrieval-strategy ablation on LongMemEval (S$^\star$). Hybrid retrieval substantially improves over single-shot summary selection; the agentic reader provides an additional gain (Table~\ref{tab:ablation}).}\label{fig:ablation-retrieval}
\end{figure}

We compare three read paths over the same maintained memory. Single-shot summary selection reaches only 41.7\%: document summaries alone miss fine-grained facts such as exact values, timestamps, and entity mentions. Adding BM25 partition retrieval (\nickname{}-H) lifts accuracy to 76.0\%, recovering most of the gap. The agentic reader (\nickname{}-A) reaches 79.3\% (Figure~\ref{fig:ablation-retrieval}) by issuing follow-up searches, inspecting local context, and combining complementary evidence spans before answering.

\subsubsection{Split-Threshold Sensitivity}
We sweep the document split threshold, defined as the minimum token count above which a topic document becomes a split candidate, on LongMemEval (S$^\star$) with the read path fixed at \nickname{}-H.

Table~\ref{tab:split-threshold} shows that the accuracy curve is asymmetric. Lowering the threshold to $\geq$3000 or $\geq$1000 costs only 1.7 and 2.0 points, even though the library expands to 762 documents at the most aggressive setting. The partitions stay topically clustered at this granularity, so over-fragmentation is recoverable: the hybrid reader still reaches most of the relevant evidence through summary plus BM25 matching. Raising the threshold to $\geq$7000 or $\geq$9000, by contrast, costs 5.3 and 6.3 points while the document count moves only modestly from 322 to 280 and then 255. The sharp accuracy drop between $\geq$5000 and $\geq$7000 therefore reflects the cost of letting documents grow past a single coherent topic rather than a count-based artefact: a small number of oversized documents accumulate content from multiple chunks and begin to mix unrelated subtopics, which degrades both topic routing during consolidation and partition-level retrieval at read time.

\begin{figure}[t]
\centering
\includegraphics[width=\columnwidth]{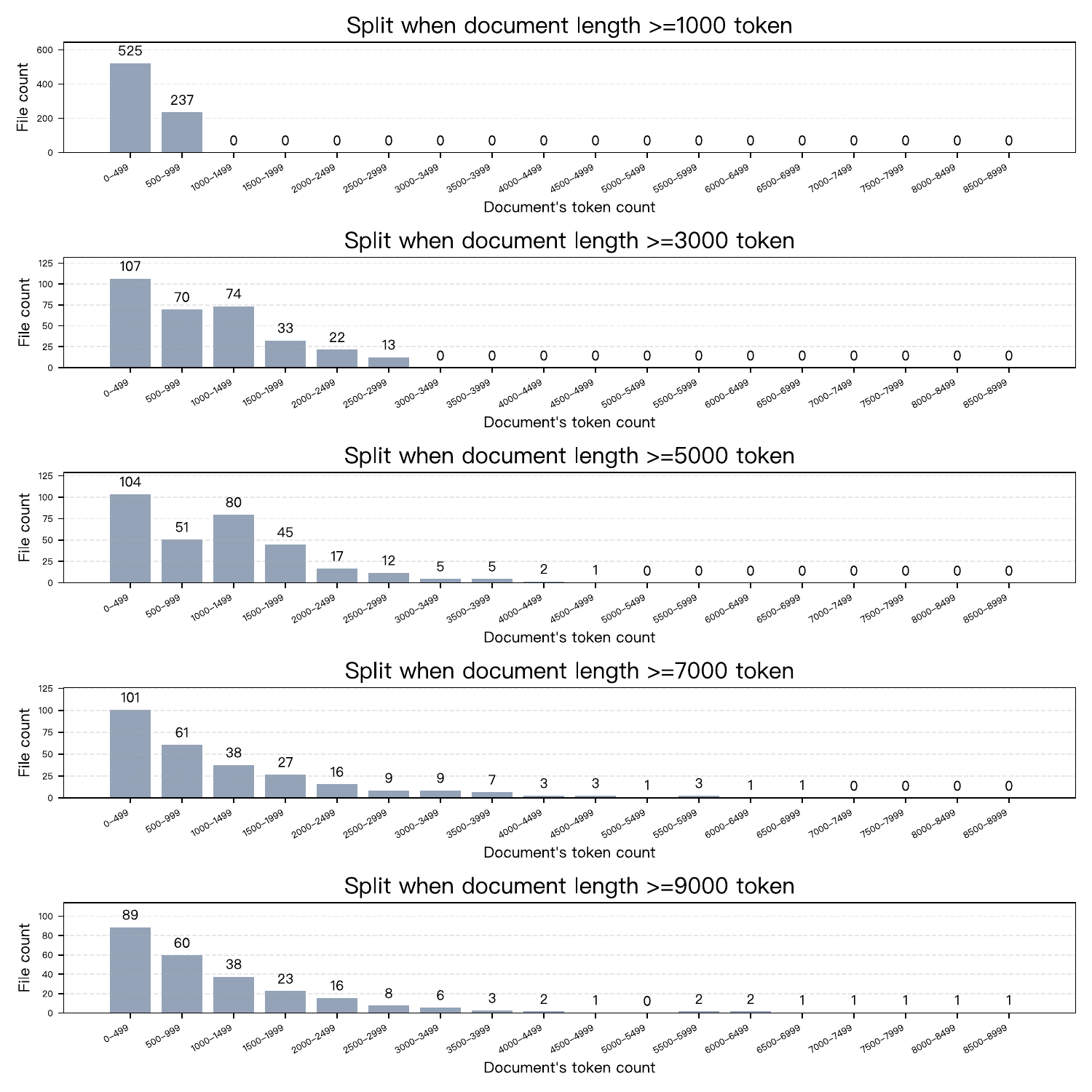}
\caption{Distribution of document token counts under different split thresholds.}\label{fig:doc-token-dist}
\end{figure}

The default $\geq$5000 setting sits just above the 4096-token chunking budget (Section~\ref{sec:setup}). With this margin, only documents that have absorbed content from more than one chunk become split candidates, so splitting acts as a safety valve for that minority rather than reshaping the bulk of the library. Figure~\ref{fig:doc-token-dist} shows the resulting token-count distributions: at $\geq$1000, nearly all documents collapse below 1000 tokens; at $\geq$9000, a long tail extends beyond 7000 tokens; at $\geq$5000, the bulk stays below 4000 tokens with only a thin tail approaching the threshold.

Overall, the ablations attribute the gains to two complementary sources. Holding the hybrid reader fixed, removing structural maintenance costs 6.7 points (76.0$\rightarrow$69.3), while holding the maintained memory fixed, upgrading from \nickname{}-H to \nickname{}-A adds 3.3 points (76.0$\rightarrow$79.3). Structural maintenance therefore contributes more than the retrieval upgrade in this setting, and neither component is sufficient on its own.

\section{Conclusion}
We presented \nickname{}, a persistent memory architecture that represents agent memory as topic-structured text documents and maintains them through buffered writing, periodic consolidation, and structural maintenance. At inference, the LLM iteratively queries memory through tool calls, keeping the memory state inspectable and editable across long-term interaction. On MemoryAgentBench, its agentic retrieval variant achieves 64.7\% overall and 81.2\% on Accurate Retrieval under our evaluation protocol, with notable gains on Factual  Recall, Test-Time Learning, and Selective Forgetting; the hybrid summary-plus-BM25 reader remains useful for long-range detailed QA\@. These results suggest that persistent agent memory quality depends on both how memory is maintained and how evidence is retrieved.

\bibliography{custom}

@misc{chhikara2025mem0,
  title         = {{Mem0}: Building Production-Ready {AI} Agents with Scalable Long-Term Memory},
  author        = {Chhikara, Prateek and Khant, Dev and Aryan, Saket and Singh, Taranjeet and Yadav, Deshraj},
  year          = {2025},
  eprint        = {2504.19413},
  archivePrefix = {arXiv},
  primaryClass  = {cs.CL},
}

@misc{fang2026lightmem,
  title         = {{LightMem}: Lightweight and Efficient Memory-Augmented Generation},
  author        = {Fang, Jizhan and Deng, Xinle and Xu, Haoming and Jiang, Ziyan and Tang, Yuqi and Xu, Ziwen and Deng, Shumin and Yao, Yunzhi and Wang, Mengru and Qiao, Shuofei and Chen, Huajun and Zhang, Ningyu},
  year          = {2025},
  eprint        = {2510.18866},
  archivePrefix = {arXiv},
  primaryClass  = {cs.CL},
  note          = {To appear in ICLR 2026},
}

@inproceedings{gutierrez2025hipporag2,
  title         = {From {RAG} to Memory: Non-Parametric Continual Learning for Large Language Models},
  author        = {Guti{\'e}rrez, Bernal Jim{\'e}nez and Shu, Yiheng and Qi, Weijian and Zhou, Sizhe and Su, Yu},
  booktitle     = {Proceedings of the 42nd International Conference on Machine Learning},
  series        = {Proceedings of Machine Learning Research},
  volume        = {267},
  pages         = {21497--21515},
  year          = {2025},
  eprint        = {2502.14802},
  archivePrefix = {arXiv},
  primaryClass  = {cs.CL},
  url           = {https://proceedings.mlr.press/v267/gutierrez25a.html},
}

@misc{hu2025memoryagentbench,
  title         = {Evaluating Memory in {LLM} Agents via Incremental Multi-Turn Interactions},
  author        = {Hu, Yuanzhe and Wang, Yu and McAuley, Julian},
  year          = {2025},
  eprint        = {2507.05257},
  archivePrefix = {arXiv},
  primaryClass  = {cs.CL},
}

@inproceedings{xu2025amem,
  title         = {{A-MEM}: Agentic Memory for {LLM} Agents},
  author        = {Xu, Wujiang and Liang, Zujie and Mei, Kai and Gao, Hang and Tan, Juntao and Zhang, Yongfeng},
  booktitle     = {Advances in Neural Information Processing Systems},
  year          = {2025},
  eprint        = {2502.12110},
  archivePrefix = {arXiv},
  primaryClass  = {cs.CL},
  url           = {https://proceedings.neurips.cc/paper_files/paper/2025/hash/19909c36f51abc4856b4560aff3d36d6-Abstract-Conference.html},
}

@inproceedings{maharana2024locomo,
  title         = {Evaluating Very Long-Term Conversational Memory of {LLM} Agents},
  author        = {Maharana, Adyasha and Lee, Dong-Ho and Tulyakov, Sergey and Bansal, Mohit and Barbieri, Francesco and Fang, Yuwei},
  booktitle     = {Proceedings of the 62nd Annual Meeting of the Association for Computational Linguistics (Volume 1: Long Papers)},
  pages         = {13851--13870},
  year          = {2024},
  eprint        = {2402.17753},
  archivePrefix = {arXiv},
  primaryClass  = {cs.CL},
  doi           = {10.18653/v1/2024.acl-long.747},
}

@inproceedings{park2023generativeagents,
  title     = {Generative Agents: Interactive Simulacra of Human Behavior},
  author    = {Park, Joon Sung and O'Brien, Joseph C. and Cai, Carrie J. and Morris, Meredith Ringel and Liang, Percy and Bernstein, Michael S.},
  booktitle = {Proceedings of the 36th Annual ACM Symposium on User Interface Software and Technology},
  year      = {2023},
  doi       = {10.1145/3586183.3606763},
}

@misc{packer2023memgpt,
  title         = {{MemGPT}: Towards {LLM}s as Operating Systems},
  author        = {Packer, Charles and Wooders, Sarah and Lin, Kevin and Fang, Vivian and Patil, Shishir G. and Stoica, Ion and Gonzalez, Joseph E.},
  year          = {2023},
  eprint        = {2310.08560},
  archivePrefix = {arXiv},
  primaryClass  = {cs.AI},
}

@misc{rasmussen2025zep,
  title         = {{Zep}: A Temporal Knowledge Graph Architecture for Agent Memory},
  author        = {Rasmussen, Preston and Paliychuk, Pavlo and Beauvais, Travis and Ryan, Jack and Chalef, Daniel},
  year          = {2025},
  eprint        = {2501.13956},
  archivePrefix = {arXiv},
  primaryClass  = {cs.CL},
}

@inproceedings{shinn2023reflexion,
  title         = {Reflexion: Language Agents with Verbal Reinforcement Learning},
  author        = {Shinn, Noah and Cassano, Federico and Gopinath, Ashwin and Narasimhan, Karthik and Yao, Shunyu},
  booktitle     = {Advances in Neural Information Processing Systems},
  volume        = {36},
  year          = {2023},
  eprint        = {2303.11366},
  archivePrefix = {arXiv},
  primaryClass  = {cs.AI},
  url           = {https://proceedings.neurips.cc/paper_files/paper/2023/hash/1b44b878bb782e6954cd888628510e90-Abstract-Conference.html},
}

@misc{shu2026remem,
  title         = {{REMem}: Reasoning with Episodic Memory in Language Agent},
  author        = {Shu, Yiheng and Jonnalagedda, Saisri Padmaja and Gao, Xiang and Guti{\'e}rrez, Bernal Jim{\'e}nez and Qi, Weijian and Das, Kamalika and Sun, Huan and Su, Yu},
  year          = {2026},
  eprint        = {2602.13530},
  archivePrefix = {arXiv},
  primaryClass  = {cs.AI},
  note          = {To appear in ICLR 2026},
}

@article{sumers2023coala,
  title         = {Cognitive Architectures for Language Agents},
  author        = {Sumers, Theodore R. and Yao, Shunyu and Narasimhan, Karthik and Griffiths, Thomas L.},
  journal       = {Transactions on Machine Learning Research},
  year          = {2024},
  eprint        = {2309.02427},
  archivePrefix = {arXiv},
  primaryClass  = {cs.AI},
  url           = {https://openreview.net/forum?id=1i6ZCvflQJ},
}

@inproceedings{wu2025longmemeval,
  title         = {{LongMemEval}: Benchmarking Chat Assistants on Long-Term Interactive Memory},
  author        = {Wu, Di and Wang, Hongwei and Yu, Wenhao and Zhang, Yuwei and Chang, Kai-Wei and Yu, Dong},
  booktitle     = {International Conference on Learning Representations},
  year          = {2025},
  eprint        = {2410.10813},
  archivePrefix = {arXiv},
  primaryClass  = {cs.CL},
  url           = {https://openreview.net/forum?id=pZiyCaVuti},
}

@inproceedings{yao2023react,
  title         = {{ReAct}: Synergizing Reasoning and Acting in Language Models},
  author        = {Yao, Shunyu and Zhao, Jeffrey and Yu, Dian and Du, Nan and Shafran, Izhak and Narasimhan, Karthik and Cao, Yuan},
  booktitle     = {International Conference on Learning Representations},
  year          = {2023},
  eprint        = {2210.03629},
  archivePrefix = {arXiv},
  primaryClass  = {cs.CL},
  url           = {https://openreview.net/forum?id=WE_vluYUL-X},
}

@inproceedings{zhong2024memorybank,
  title     = {{MemoryBank}: Enhancing Large Language Models with Long-Term Memory},
  author    = {Zhong, Wanjun and Guo, Lianghong and Gao, Qiqi and Ye, He and Wang, Yanlin},
  booktitle = {Proceedings of the AAAI Conference on Artificial Intelligence},
  volume    = {38},
  pages     = {19724--19731},
  year      = {2024},
  doi       = {10.1609/aaai.v38i17.29946},
}

@inproceedings{qian2025memorag,
  title     = {{MemoRAG}: Boosting Long Context Processing with Global Memory-Enhanced Retrieval Augmentation},
  author    = {Qian, Hongjin and Liu, Zheng and Zhang, Peitian and Mao, Kelong and Lian, Defu and Dou, Zhicheng and Huang, Tiejun},
  booktitle = {Proceedings of the ACM on Web Conference 2025},
  pages     = {2366--2377},
  year      = {2025},
  doi       = {10.1145/3696410.3714805},
}

@inproceedings{sarthi2024raptor,
  title         = {{RAPTOR}: Recursive Abstractive Processing for Tree-Organized Retrieval},
  author        = {Sarthi, Parth and Abdullah, Salman and Tuli, Aditi and Khanna, Shubh and Goldie, Anna and Manning, Christopher},
  booktitle     = {International Conference on Learning Representations},
  year          = {2024},
  eprint        = {2401.18059},
  archivePrefix = {arXiv},
  primaryClass  = {cs.CL},
  url           = {https://openreview.net/forum?id=GN921JHCRw},
}

@inproceedings{tan2025prospect,
  title         = {In Prospect and Retrospect: Reflective Memory Management for Long-term Personalized Dialogue Agents},
  author        = {Tan, Zhen and Yan, Jun and Hsu, I-Hung and Han, Rujun and Wang, Zifeng and Le, Long T. and Song, Yiwen and Chen, Yanfei and Palangi, Hamid and Lee, George and Iyer, Anand Rajan and Chen, Tianlong and Liu, Huan and Lee, Chen-Yu and Pfister, Tomas},
  booktitle     = {Proceedings of the 63rd Annual Meeting of the Association for Computational Linguistics (Volume 1: Long Papers)},
  pages         = {8416--8439},
  year          = {2025},
  doi           = {10.18653/v1/2025.acl-long.413},
}

@inproceedings{kang2025memory,
  title         = {Memory {OS} of {AI} Agent},
  author        = {Kang, Jiazheng and Ji, Mingming and Zhao, Zhe and Bai, Ting},
  booktitle     = {Proceedings of the 2025 Conference on Empirical Methods in Natural Language Processing},
  pages         = {25961--25970},
  year          = {2025},
  doi           = {10.18653/v1/2025.emnlp-main.1318},
}

\clearpage
\onecolumn
\raggedbottom

\appendix

\section{Prompt Templates and Algorithms}
The released reference implementation contains functionally equivalent prompts in Chinese. For reproducibility, this appendix provides normalized English templates that preserve the same control logic, metadata constraints, and output schemas used by the system.

\subsection{Normalized English Prompt Templates}\label{appendix:prompts}

\paragraph{Shared prompt invariants.} All write-side prompts follow four invariants. First, they must remain fully faithful to the source text and may reorganize content but must not invent new facts. Second, they must preserve entry-level metadata such as \texttt{seq}, \texttt{time}, \texttt{source}, and any additional fields such as \texttt{label}. Third, they must resolve exact duplicates and contradictions through explicit recency rules rather than free-form summarization. Fourth, they must return machine-parseable output only, with no explanatory prose outside the requested format.

\begin{widepromptbox}[Prompt~{A}: Memory Extraction]
You are a memory curator for a long-term interactive agent.\\
Extract only durable, reusable memory candidates from the interaction.\\[4pt]
\textbf{Retain information such as:}\\
-- user facts, preferences, plans, constraints, commitments, and corrections\\
-- important tool or environment observations that may matter later\\
-- assistant restatements only when they explicitly confirm user information; mark such items with source=AI\\
-- explicit metadata fields already present in the text, such as label=\ldots\\[4pt]
\textbf{Do not retain:}\\
-- greetings, small talk, generic acknowledgments, or filler politeness\\
-- transient reasoning, unsupported guesses, or speculative interpretation\\
-- duplicate paraphrases that do not change the underlying state\\[4pt]
\textbf{Hard constraints:}\\
-- stay fully faithful to the source text\\
-- do not add explanations, comments, or fact checking\\
-- preserve metadata fields exactly when they already appear in the input\\
-- keep the literal placeholder @@SEQ@@ unchanged\\[4pt]
\textbf{Output format:}\\
1. Return Markdown only.\\
2. Begin with:\\
\quad\texttt{---}\\
\quad\texttt{summary: <topic keywords; concise factual summary, <= \{summary\_length\} tokens>}\\
\quad\texttt{---}\\
3. Organize the body with first-level headings.\\
4. Encode each memory item as:\\
\quad\texttt{- <seq=@@SEQ@@,time=TIMESTAMP[,label=\ldots][,source=AI]> fact}\\
5. If no durable memory is present, return a minimal empty-memory document.
\end{widepromptbox}

\begin{widepromptbox}[Prompt~{B}: CURRENT Rewrite]
You will receive the append-only CURRENT document.\\
Rewrite it into a clean, topic-structured Markdown draft.\\[4pt]
\textbf{Requirements:}\\
-- group semantically related items under first-level headings\\
-- preserve every seq / time / source / label field exactly\\
-- remove exact duplicates; if duplicates differ only by seq, keep the newer one\\
-- resolve contradictory facts by keeping the more recent item\\
-- if the same fact appears once as user input and once as source=AI, prefer the user-sourced version\\
-- do not introduce new facts, commentary, or interpretation\\[4pt]
Return Markdown only, including YAML frontmatter with an updated summary.
\end{widepromptbox}

\begin{widepromptbox}[Prompt~{C}: Update Planning]
You will receive:\\
1. \texttt{NEW\_CONTENT}: a rewritten Markdown draft\\
2. \texttt{DOCS}: the current document library as (id, summary) pairs\\[4pt]
\textbf{Task:}\\
-- decide which existing documents should be updated\\
-- decide which content should become new topic documents\\[4pt]
\textbf{Planning rules:}\\
-- preserve seq / time / source / label fields exactly\\
-- never invent document ids; updates must target ids already listed in DOCS\\
-- split content by topic when necessary\\
-- keep each produced document within the token budget \{markdown\_length\}\\
-- ensure no content overlap across outputs\\[4pt]
\textbf{Return JSON only:}\\
\{\\
\quad\jsonquote{}updates\jsonquote{}: [\{\jsonquote{}id\jsonquote{}: string, \jsonquote{}new\_content\jsonquote{}: string\}],\\
\quad\jsonquote{}new\_docs\jsonquote{}: [\{\jsonquote{}title\jsonquote{}: string, \jsonquote{}content\jsonquote{}: string\}]\\
\}
\end{widepromptbox}

\begin{widepromptbox}[Prompt~{D}: Document Rewrite]
You will receive:\\
1. \texttt{OLD\_DOC}: an existing Markdown document\\
2. \texttt{DELTA}: new content that should be merged into it\\[4pt]
\textbf{Task:}\\
-- produce the updated full document\\
-- reorganize headings when needed\\
-- preserve all entry metadata exactly\\
-- deduplicate repeated facts\\
-- resolve contradictions by recency, again preferring user-sourced facts when two otherwise identical items differ only by source\\[4pt]
Return Markdown only, including YAML frontmatter with a refreshed summary.
\end{widepromptbox}

\begin{widepromptbox}[Prompt~{E}: Agentic Retrieval]
You will receive:\\
1.\ a user query\\
2.\ the document list as (id, summary) pairs\\[4pt]
You are a memory-search agent. Iteratively inspect the topic document library before stopping. You may either:\\
-- call tools to search the corpus, inspect a single document, browse more document ids, or read bounded line ranges\\
-- or finish once you have enough evidence\\[4pt]
Prefer using broad lexical search and exact-pattern search early, then use document-local inspection to verify hits and expand context. For aggregation queries, continue until the main matching evidence is covered. For update or temporal questions, pay attention to seq, time, and source fields.\\[4pt]
\textbf{Return JSON only:}\\
\{\\
\quad\jsonquote{}done\jsonquote{}: boolean,\\
\quad\jsonquote{}tool\_calls\jsonquote{}: [\{\jsonquote{}name\jsonquote{}: string, \jsonquote{}arguments\jsonquote{}: object\}],\\
\quad\jsonquote{}relevant\_snippets\jsonquote{}: [\{\jsonquote{}doc\_id\jsonquote{}: string, \jsonquote{}start\_line\jsonquote{}: int, \jsonquote{}end\_line\jsonquote{}: int\}],\\
\quad\jsonquote{}relevant\_docs\jsonquote{}: [string, \ldots]\\
\}
\end{widepromptbox}

\paragraph{Merge-related prompts.} The implementation also uses two auxiliary prompts during scheduled consolidation: one selects non-overlapping groups of small, topically similar documents for merging, and the other rewrites the selected group into a single normalized document. Both reuse the same invariants as above: exact metadata preservation, no invented facts, deduplication by recency, and Markdown-only output.

\FloatBarrier%

\subsection{Strategy Selection and Design Rationale}\label{appendix:rationale}
The architecture deliberately prioritizes note-like structured text over a purely discrete vector store or a graph-only memory layer.

\paragraph{Why not rely only on discrete vector memory?} Discrete entries are easy to append and index, but they couple retrieval quality tightly to fragment granularity. As the topic document library grows, topic-related evidence becomes scattered and top-$k$ retrieval struggles to recover complete evidence sets.

\paragraph{Why not rely only on a knowledge graph?} Knowledge graphs are powerful for representing explicit relational structure, but many interaction memories do not naturally map to graph edges. They also increase modeling and maintenance cost, especially when temporal revision and evidence provenance matter.

\paragraph{Why use a buffer plus consolidation?} The buffer allows append-only high-frequency writes during interaction, while consolidation amortizes the cost of organizing, revising, and redistributing memory. This separation mirrors human note-taking and makes the write path predictable.

\subsection{Algorithmic Pseudocode}\label{appendix:algorithms}

\begin{algorithm}[H]
\caption{Memory Writing and Consolidation}\label{alg:write-consolidate}
\footnotesize
\begin{algorithmic}[1]
\Procedure{WriteAndConsolidate}{$z_t, C, D, \textit{cfg}$}
    \State{} $M \gets \Call{Extract}{z_t}$
    \If{$M = \emptyset$}
        \State{} \Return{} $C, D$
    \EndIf%
    \If{$C$ does not exist}
        \State{} $C \gets \Call{CreateCurrent}{M}$
    \Else%
        \State{} $C \gets \Call{AppendToCurrent}{C, M}$
    \EndIf%
    \If{$\Call{Tokens}{C} \leq \textit{cfg}.\text{current\_threshold}$}
        \State{} \Return{} $C, D$
    \EndIf%
    \State{} $R \gets \Call{RewriteCurrent}{C}$
    \State{} $S_D \gets \Call{Summaries}{D \setminus \{\textsc{current}\}}$
    \State{} $P \gets \Call{SafePlanUpdate}{R,\; S_D,\; \textit{cfg}}$
    \ForAll{update $u \in P.\text{updates}$}
        \State{} $D[u.\text{id}] \gets \Call{RewriteDoc}{D[u.\text{id}],\; u.\text{new\_content}}$
    \EndFor%
    \ForAll{new document $n \in P.\text{new\_docs}$}
        \State{} $D \gets D \cup \{\Call{RewriteDoc}{\varnothing,\; n.\text{content}}\}$
    \EndFor%
    \State{} $\Call{Clear}{\textsc{current}}$
    \State{} $\Call{IncrementCurrentEpoch}{}$
    \If{$\Call{MergeMaintenanceEnabled}{\textit{cfg}}$}
        \State{} $G_{\text{small}} \gets \Call{SmallSimilarDocs}{D}$
        \State{} $G \gets \Call{SelectMergeGroups}{G_{\text{small}}}$
        \ForAll{group $g \in G$}
            \State{} $D \gets \Call{ReplaceGroupWithMergedDoc}{D, g}$
        \EndFor%
    \EndIf%
    \State{} $\Call{RefreshLibraryMetadata}{D}$
    \State{} \Return{} $C, D$
\EndProcedure%
\end{algorithmic}
\end{algorithm}

\begin{algorithm}[H]
\caption{Split-Aware Update Planning}\label{alg:safe-plan}
\footnotesize
\begin{algorithmic}[1]
\Procedure{SafePlanUpdate}{$\textit{content}, \textit{summaries}, \textit{cfg}$}
    \If{$\Call{Tokens}{\textit{content}} > \textit{cfg}.\text{plan\_split\_threshold}$}
        \State{} $\textit{chunks} \gets \Call{SplitByHeading}{\textit{content},\; \textit{cfg}.\text{markdown\_length}}$
        \If{$|\textit{chunks}| = 1$}
            \State{} $\textit{chunks} \gets \Call{RecursiveFallbackSplit}{\textit{content},\; \textit{cfg}.\text{plan\_split\_threshold}}$
        \EndIf%
    \Else%
        \State{} $\textit{chunks} \gets \{\textit{content}\}$
    \EndIf%
    \State{} $\textit{plans} \gets [\,]$
    \ForAll{$\textit{chunk} \in \textit{chunks}$}
        \State{} $\textit{plan} \gets \Call{PlanUpdate}{\textit{chunk},\; \textit{summaries}}$
        \If{$\textit{plan}$ is not valid JSON}
            \State{} $\textit{sub} \gets \Call{RecursiveFallbackSplit}{\textit{chunk},\; \textit{cfg}.\text{plan\_split\_threshold}}$
            \State{} $\textit{subplans} \gets \Call{SafePlanUpdate}{\textit{sub}, \textit{summaries}, \textit{cfg}}$
            \State{} $\textit{plan} \gets \Call{MergePlans}{\textit{subplans}}$
        \EndIf%
        \State{} $\textit{plans} \gets \textit{plans} \cup \{\textit{plan}\}$
    \EndFor%
    \State{} \Return{} $\Call{MergePlans}{\textit{plans}}$
\EndProcedure%
\end{algorithmic}
\end{algorithm}

\begin{algorithm}[H]
\caption{Agentic Memory Retrieval}\label{alg:retrieval}
\footnotesize
\begin{algorithmic}[1]
\Procedure{RetrieveAndAnswer}{$q_t, C, D, \textit{cfg}$}
    \State{} $S \gets \Call{LibrarySummaries}{D}$
    \State{} $H \gets \{(\text{query}=q_t,\; \text{catalog}=S)\}$
    \For{$i \gets 1$ \textbf{to} $\textit{cfg}.\text{agentic\_max\_iterations}$}
        \State{} $o \gets \Call{AgentStep}{H}$
        \If{$o.\text{done}$}
            \State{} $Z \gets \Call{MaterializeSelections}{o.\text{snippets},\; o.\text{docs},\; D}$
            \State{} \textbf{break}
        \EndIf%
        \State{} $T \gets \Call{ExecuteTools}{o.\text{tool\_calls},\; D,\; \textit{cfg}}$
        \State{} $H \gets H \cup \{T\}$
    \EndFor%
    \If{$Z = \emptyset$ \textbf{or} $\Call{Tokens}{Z} < \textit{cfg}.\text{min\_retrieval\_tokens}$}
        \State{} $P_D \gets \Call{HeadingPartitions}{D}$
        \State{} $Z \gets Z \cup \Call{BM25Fallback}{q_t,\; P_D,\; \textit{cfg}.\text{fallback\_topk}}$
    \EndIf%
    \State{} $E \gets Z \cup \Call{Recent}{C}$
    \State{} $a_t \gets \Call{LLMAnswer}{q_t,\; E}$
    \State{} \Return{} $a_t$
\EndProcedure%
\end{algorithmic}
\end{algorithm}

\end{document}